\documentclass{article}
\usepackage{ijcai21}

\usepackage{times}

\usepackage{soul}
\usepackage{url}
\usepackage[hidelinks]{hyperref}
\usepackage[utf8]{inputenc}
\usepackage[small]{caption}
\usepackage{graphicx}
\usepackage{amsmath}
\usepackage{booktabs}
\usepackage{times}
\usepackage{epsfig}
\usepackage{graphicx}
\usepackage{amsmath}
\usepackage{amssymb}
\usepackage{booktabs}
\usepackage{dsfont} 
\usepackage{color}
\usepackage{pifont}
\usepackage{graphics}
\usepackage{array}
\usepackage{multirow}
\usepackage{dblfloatfix}
\usepackage{enumitem}
\title{Context-Conditional Adaptation for Recognizing Unseen Classes in Unseen Domains}

\author{
Puneet Mangla \thanks{Equal Contribution} $^1$\and
Shivam Chandhok \footnotemark[1] $^{2}$\and\\
Vineeth N Balasubramanian$^1$\And
Fahad Shahbaz Khan$^2$
\affiliations
$^1$Department of Computer Science, Indian Institute of Technology, Hyderabad\\
$^2$
Mohamed bin Zayed University of Artificial Intelligence\\
\emails
pmangla261@gmail.com,
shivam.chandhok@mbzuai.ac.ae, vineethnb@cse.iith.ac.in, fahad.khan@mbzuai.ac.ae
}

\begin{document}

\maketitle

\begin{abstract}
 Recent progress towards designing models that can generalize to unseen domains (i.e domain generalization) or unseen classes (i.e zero-shot learning) has embarked interest towards building models that can tackle both domain-shift and semantic shift simultaneously (i.e zero-shot domain generalization). For models to generalize to unseen classes in unseen domains, it is crucial to learn feature representation that preserves class-level (domain-invariant) as well as domain-specific information. Motivated from the success of generative zero-shot approaches, we propose a feature generative framework integrated with a \underline{CO}ntext \underline{CO}nditional \underline{A}daptive (COCOA) Batch-Normalization to seamlessly integrate class-level semantic and domain-specific information. The generated visual features better capture the underlying data distribution enabling us to generalize to unseen classes and domains at test-time. We thoroughly evaluate and analyse our approach on established large-scale benchmark - DomainNet and demonstrate promising performance over baselines and state-of-art methods. 
\end{abstract}

\section{Introduction}

The dependence of deep learning models on large amounts of data and supervision creates a bottleneck and hinders their utilization in practical scenarios. 
There is thus a need to equip deep learning models with the ability to generalize to unseen domains or classes at test-time using data from other related domains or classes (where data is abundant).


There has been great interest and corresponding efforts in recent years towards tackling \textit{domain shift} (a.k.a Domain Generalization) \cite{basicdg1,basicdg2,basicdg3,basicdg4,MTAE,DAFL,jigsaw,metadg1,metadg2} and \textit{semantic shift} (a.k.a Zero-Shot Learning) \cite{vis2semantic1,devise,semantic2vis1,semantic2vis2,semantic2vis3} separately.
However, applications in the real world do not guarantee that only one of them will occur -- there is a need to make systems robust to the domain and semantic shifts simultaneously. The problem of tackling domain shift and semantic shift together, as considered herein, can be grouped as the Zero-Shot Domain Generalization (which we call ZSLDG) problem setting \cite{dgzsl,ZeroShotDG}. In ZSLDG, the model has access to a set of seen classes from multiple source domains and has to generalize to unseen classes in unseen target domains at test time. Note that this problem of recognizing unseen classes in unseen domains (ZSLDG) is much more challenging than tackling zero-shot learning (ZSL) or domain generalization (DG) separately \cite{dgzsl} and growingly relevant as deep learning models get reused across domains. 
A recent approach, CuMix \cite{dgzsl}, proposed a methodology that mixes up multiple source domains and categories available during training to simulate semantic and domain shift at test time. This work also established a benchmark dataset, DomainNet, for the ZSLDG problem setting with an evaluation protocol, which we follow in this work for a fair comparison. \\
\noindent Feature generation methods \cite{zslgen0,zslgen2,zslgen3,zslgen4,zslgen5,zslgen6,zslgen7,tfvaegan,taco,normflows} have recently shown significant promise in traditional zero-shot or few-shot learning settings and are among the state-of-the-art today for such problems. However, these approaches assume that the source domains (during training) and the target domain (during testing) are the same and thus aim to generate features that only address semantic shift. They rely on the assumption that the visual-semantic relationship learned during training will generalize to unseen classes at test-time. However, there is no guarantee that this holds for images in novel domains unseen during training \cite{dgzsl}. Thus, when used for addressing ZSLDG, the aforementioned methods lead to suboptimal performance.\\
To tackle domain shift at test-time, it is important to generate feature representations that encode both class level (domain-invariant) and domain-specific information  \cite{BalanceSpecInv,BNE2}. Thus we conjecture that generating features that are consistent with only class-level semantics is not sufficient for addressing ZSLDG. However, encoding domain-specific information along with class information in generated features is not straightforward \cite{dgzsl}.
\begin{figure*}[t]

\includegraphics[width=\linewidth]{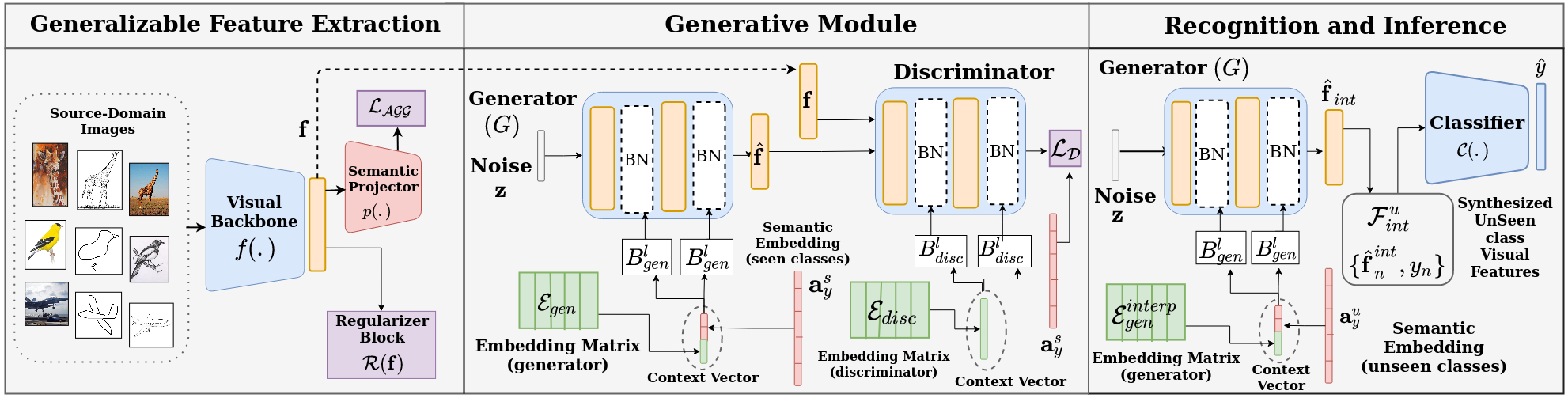}
\vspace{-20pt}
\caption{\footnotesize The proposed pipeline consists of three stages: (1) Generalizable Feature Extraction; (2) Generative Module; and (3) Recognition and Inference. The first stage trains a visual backbone $f(.)$ to extract visual feature $\mathbf{f}$ that encodes discriminative class-level (domain-invariant) and domain-specific information. The second stage learns a generative model $G$ which uses \underline{CO}ntext \underline{CO}nditioned \underline{A}daptive (COCOA) batch-normalization layers to fuse and integrate semantic and domain-specific information into the generated features $\hat{\mathbf{f}}$. Lastly, the third stage generates visual features for unseen classes across domains and trains a softmax classifier }
\vspace{-17pt}
\label{fig:methodology}
\end{figure*}
Drawing inspiration from these observations, we propose a unified generative framework that uses \underline{CO}ntext \underline{CO}nditional \underline{A}daptive (COCOA) Batch-Normalization to seamlessly integrate semantic and domain information to generate visual features of unseen classes in unseen domains. Depending on the setting, ``\textit{context}'' can qualify as domain (for DG), semantics (for ZSL), or both (in case of ZSLDG). Since this work primarily deals with the ZSLDG setting, 'context' in this work refers to both domain and semantic information. We perform experiments and analysis on standard ZSLDG benchmark: DomainNet and demonstrate that our proposed methodology provides state-of-the-art performance for the ZSLDG setting and is able to encode both semantic and domain-specific information to generate features, thus capturing the given context better.
To the best of our knowledge, this is the first generative approach to address the ZSLDG problem setting.

\section{COCOA: Proposed Methodology}
Our goal is to train a classifier $\mathcal{C}$ which can tackle domain-shift as well as semantic shift simultaneously and recognize unseen classes in unseen domains at test-time. Let  $S^{Tr}=\{ (\mathbf{x},  y,  \mathbf{a}_y^{s}, d )| \mathbf{x} \in \mathcal{X},  y \in \mathcal{Y}^s,  \mathbf{a}_y^{s} \in \mathcal{A},  d \in \mathcal{D}^s\}$ denote the training set, where $\mathbf{x}$ is a seen class image in the visual space ($\mathcal{X}$) with corresponding label $y$ from a set of seen class labels $\mathcal{Y}^s$. $\mathbf{a}_y^{s}$ denotes the class-specific semantic representation for seen classes. We assume access to domain labels $d$ for a set of source domains $\mathcal{D}^s$ with cardinality $K$. 
The test-set is denoted by $S^{Ts}=\{ (\mathbf{x},  y,  \mathbf{a}_y^{u}, d )| \mathbf{x} \in \mathcal{X},  y \in \mathcal{Y}^u,  \mathbf{a}_y^{u} \in \mathcal{A},  d \in \mathcal{D}^u\}$ where $\mathcal{Y}^u$ is the set of labels for unseen classes and $\mathcal{D}^u$ represents the set of unseen target domains. Note that standard zero-shot setting (tackles only semantic-shift) complies with the condition $\mathcal{Y}^s \cap \mathcal{Y}^u \equiv \emptyset$ and $\mathcal{D}^s\equiv\mathcal{D}^u$. The standard DG setting (tackles only domain-shift), on the other hand, works under the condition $\mathcal{Y}^s\equiv\mathcal{Y}^u $ and $\mathcal{D}^s \cap \mathcal{D}^u \equiv \emptyset$. 
In this work, our goal is to address the challenging ZSLDG setting where $\mathcal{Y}^s\cap\mathcal{Y}^u \equiv \emptyset$ and $\mathcal{D}^s\cap\mathcal{D}^u \equiv \emptyset$. 

Our overall framework to address ZSLDG is summarized in Fig \ref{fig:methodology}. We employ a three-stage pipeline, which we describe below.

\subsection{Generalizable Feature Extraction}
\label{feat_extract}
We extract visual features $\mathbf{f}$ that encode discriminative class-level cues (domain-invariant) as well as domain-specific information by training a visual encoder $f(.)$, which is then used to train a semantic projector $p(.)$. Both these modules are trained with images from all source domains available. 
The visual encoder and the semantic projector are trained to minimize the following loss:
\vspace{-3pt}
\begin{align}
\mathcal{L}_{AGG} = \mathbb E_{(\mathbf{x},y) \sim S^{Tr} } \left [ \mathcal{L}_{CE}(\mathbf{a}^T p(f(\mathbf{x})), y) \right ] +  \mathcal{R}(\mathbf{f})
\vspace{-3pt} 
\label{agg_obj}
\end{align}
where $\mathcal{L}_{CE}$ is standard cross-entropy loss and $\mathbf{a} = [\mathbf{a}^s_1, ..., \mathbf{a}^s_{\vert \mathcal{Y}^s \vert}]$. $\mathcal{R}(\mathbf{f})$ denotes the regularization loss term.
which is used to further improve the visual features and enhance their generalizability by regulating the amount of class-level semantic and domain-specific information in the features $\mathbf{f}$. The regularizer block primarily uses rotation prediction (viz. providing a rotated image as input, and predicting rotation angle). 
\vspace{-4pt}

\subsection{Generative Module} \label{gen}
\noindent 
We learn a generative model which comprises of a generator $G:\mathcal{Z}\times\mathcal{A}\times\mathcal{D} \to \mathcal{F}$ and a projection  discriminator \cite{miyato2018cgans} (which uses a projection layer to incorporate conditional information into the discriminator) $D:\mathcal{F}\times\mathcal{A}\times\mathcal{D}\to \mathbb R$ as shown in Fig \ref{fig:methodology}. We represent this discriminator as $D = D_l \circ D_f$ ($\circ$ denotes composition) where $D_f$ is discriminator’s last feature layer and $D_l$ is the final linear layer.
Both the generator and the projection discriminator are conditioned through a Context Conditional Adaptive (COCOA) Batch-Normalization module, which provides class-specific and domain-specific modulation at various stages during the generation process.
Modulating the feature information by such semantic and domain-specific embeddings helps to integrate respective information into the features, thereby enabling better generalization. 
We use the available (seen) semantic attributes $\mathbf{a}^s_{y}$ that capture class-level characteristics, for encoding semantic information. However, such a representation that encodes domain information is not present for  individual source domains. We hence define learnable (randomly initialized and optimized during training) domain embedding matrices $\mathcal{E}_{gen}=\{\mathbf{e}_{1}^{gen}, \mathbf{e}_{2}^{gen}, \mathbf{e}_{3}^{gen}..., .\mathbf{e}_{K}^{gen}\} $ and $\mathcal{E}_{disc}=\{\mathbf{e}_{1}^{disc}, \mathbf{e}_{2}^{disc}, \mathbf{e}_{3}^{disc}....\mathbf{e}_{K}^{disc}\} $ for the generator and the discriminator respectively.

The generator $G$ takes in noise $\textbf{z} \in \mathcal{Z}$ and a context vector $\textbf{c}$, and outputs visual features $\hat{\mathbf{f}} \in \mathcal{F} $ . The context vector $\textbf{c}$, which is the concatenation of class-level semantic attribute $\mathbf{a}^s_{y}$ and domain-specific embedding $\mathbf{e}_{i}^{gen}$, is provided as input to a BatchNorm estimator network $B^l_{gen}:\mathcal{A} \times \mathcal{D} \to \mathcal{R}^{2 \times h}$ ($h$ is the dimension of layer $l$ activation vectors) which outputs  batchnorm paramaters, $\gamma^l_{gen}$ and $\beta^l_{gen}$ for the $l$-th layer. Similarly, the discriminator's feature extractor $D_f(.)$ has a separate BatchNorm estimator network $B^l_{disc}:\mathcal{D} \to \mathcal{R}^{2 \times h}$ to enable domain-specific context modulation of its batchnorm parameters, $\gamma^l_{disc} ,\beta^l_{disc}$ as shown in Fig \ref{fig:methodology}.

Formally, let $\mathbf{f}^l_{gen}$ and $\mathbf{f}^l_{disc}$ denote feature activations belonging to domain $d$ and semantic attribute $\mathbf{a}^s_y$, at $l$-th layer of generator $G$ and discriminator $D$ respectively. We modulate $\mathbf{f}^l_{gen}$ and $\mathbf{f}^l_{disc}$ individually using Context Conditional Adaptive Batch-Normalization as follows:
\begin{equation}
\small
\begin{split}
& \gamma^l_{gen} ,\beta^l_{gen} \gets B^l_{gen}(\textbf{c}) \; , \; \mathbf{f}^{l+1}_{gen} \gets \gamma^l_{gen} \cdot \frac{\mathbf{f}^l_{gen} - \mu^l_{gen}}{\sqrt{(\sigma^l_{gen})^2 + \epsilon}} + \beta^l_{gen} \\
& \text{where} \; \; \textbf{c} = [\mathbf{a}^s_{y},\mathbf{e}_{d}^{gen}] \\
&  \gamma^l_{disc} ,\beta^l_{disc} \gets B^l_{disc}(\mathbf{e}_{d}^{disc}) \; , \; \mathbf{f}^{l+1}_{disc} \gets \gamma^l_{disc} \cdot \frac{\mathbf{f}^l_{disc} - \mu^l_{disc}}{\sqrt{(\sigma^l_{disc})^2 + \epsilon}} + \beta^l_{disc}
\end{split}
    \label{eqn:prior}
\end{equation}
Here, $(\mu^l_{gen}, (\sigma^l_{gen})^2)$ and $(\mu^l_{disc}, (\sigma^l_{disc})^2)$ are the mean and variance of activations of the mini-batch (also used to update running statistics) containing $\mathbf{f}^l_{gen}$ and $\mathbf{f}^l_{disc}$ respectively. $\textbf{c}$ denotes the context vector composed of semantic and domain embeddings and $[\cdot,\cdot]$ denotes the concatenation operation.

Finally, the generator and discriminator are trained to optimize the adversarial loss given by:
\begin{align}
\small
\begin{split}
 \mathcal{L}_D & = \mathbb{E}_{(\mathbf{x}, \mathbf{a}^s_y, d) \sim S^{Tr}} \left [ \text{max}(0, 1 - D(f(\mathbf{x}), \mathbf{a}^s_y, \mathbf{e}^{disc}_d)) \right ]  \\
& + \mathbb{E}_{y \sim \mathcal{Y}^s, d \sim \mathcal{D}^s, \textbf{z} \sim \mathcal{Z}} \left [ \text{max}(0, 1 + D(\hat{\mathbf{f}}, \mathbf{a}^s_y, \mathbf{e}^{disc}_d)) \right ]     
\end{split}
\end{align}
\begin{align}
\small
\begin{split}
\mathcal{L}_G & = \mathbb{E}_{y \sim \mathcal{Y}^s, d \sim \mathcal{D}^s, \textbf{z} \sim \mathcal{Z}} [ -D(\hat{\mathbf{f}}, \mathbf{a}^s_y, \mathbf{e}^{disc}_d) \\
& + \lambda_G \cdot  \mathcal{L}_{CE}(\mathbf{a}^Tp(\hat{\mathbf{f}}), y) ] 
\end{split}
\end{align}
where $D(\mathbf{f},\mathbf{a},\mathbf{e}) = \mathbf{a}^T D_f(\mathbf{f},\mathbf{e}) + D_l( D_f(\mathbf{f},\mathbf{e}))$ is the projection term ,  $\hat{\mathbf{f}}=G(\textbf{z}, \textbf{c})$ denotes generated feature. The second term, $\mathcal{L}_{CE}(\mathbf{a}^Tp(\hat{\mathbf{f}}), y)$ in $\mathcal{L}_G$ ensures that the generated features have discriminative properties ($\lambda_G$ is a hyperparameter). $p(.)$ is the semantic projector that was trained in the previous stage.  

\subsection{Recognition and Inference}
\label{inf}
   We freeze the generator and aim to synthesize visual features for unseen classes across different domains, which are then used to train classifier $\mathcal{C}$. To this end, we concatenate the domain embeddings $\mathbf{e}_{i}^{gen}, (i \in  \mathcal{D}^s$) of the source domains from matrix $\mathcal{E}_{gen}$ (learned end-to-end with the generative model) and the semantic attributes/representations of unseen classes $\mathbf{a}^u_{y}$ to get the context vector $\textbf{c}$, which in turn is input to the trained batchnorm predictor network $B^l_{gen}$. The output batchnorm parameters $(\gamma_{gen}^l, \beta_{gen}^l)$ are used in the batchnorm layers of the pre-trained generator to generate features $\hat{\mathbf{f}}$. This enables us to generate features that are consistent with unseen class semantics and also encode domain information from individual source domains in the training set. After obtaining batch-norm parameters, the new set of unseen class features 
are generated as follows:
\vspace{-4pt}
\begin{equation}
\small
\label{no-mix-inference}
\begin{split}
 & \hat{\mathbf{f}}_n  = G(\textbf{z}_n, \textbf{c}) \\ 
 & \text{where} \   \textbf{c} = [\mathbf{a}^u_{y_n}, \mathbf{e}_{n}^{gen}], \textbf{z}_n \sim \mathcal{Z}, y_n \sim \mathcal{Y}^u, \mathbf{e}_{n}^{gen} \sim \mathcal{E}_{gen}   
\vspace{-4pt}
\end{split}
\end{equation}

To improve generalization to new domains at test-time and make the classifier domain-agnostic, we synthesize embeddings of newer domains by interpolating the learned embeddings of the source domains via a mix-up operation.
\begin{equation}
\small
\label{eq:wgan}
\mathcal{E}_{interp}^{gen} = \lambda \cdot \mathbf{e}_{i}^{gen} + (1-\lambda) \cdot \mathbf{e}_{j}^{gen} \  \text{ where } \ i, j \sim \mathcal{D}^s, \lambda \sim \mathcal{U}[0,1]
\vspace{-1pt}                 
\end{equation}
where $\mathbf{e}_{i}^{gen}$ and $\mathbf{e}_{j}^{gen}$ refer to the domain embeddings of $i$-th and $j$-th source domains.
The unseen class features generated using interpolated domain embeddings
$\mathcal{F}^u_{int} = \{ (\hat{\mathbf{f}}^n_{int}, y_n) \}_{n=1}^N$ 
are generated as follows:
\vspace{-4pt}
\begin{equation}
\small
\label{mix-inference}
\begin{split}
& \hat{\mathbf{f}}^n_{int}  = G(\textbf{z}_n, \textbf{c}_{interp.} ) \; \; \text{where} \; \; \textbf{c}_{interp.} = [\mathbf{a}^u_{y_n}, \mathbf{e}_{interp.}^{gen}],
\\ & \textbf{z}_n \sim \mathcal{Z}, y_n \sim \mathcal{Y}^u, \mathbf{e}_{interp.}^{gen} \sim \mathcal{E}_{interp}^{gen} 
\end{split}
\vspace{-4pt}
\end{equation}
Next, we train a MLP softmax classifier, $\mathcal{C}(.)$ on the generated multi-domain unseen class features dataset $\mathcal{F}^u_{int}$  by minimizing: $\mathcal{L}_{CLS} = \mathbb E_{(\hat{\mathbf{f}}_{int}, y) \sim \mathcal{F}^u_{int}}[ \mathcal{L}_{CE}(\mathcal{C}(\hat{\mathbf{f}}_{int}), y)]$


\noindent \textbf{Classifying image at test time.} At test-time, given a test image $\mathbf{x}_{test}$, we first pass it through the visual encoder $f(.)$ to get the discriminative feature representation $\mathbf{f}_{test} = f(\mathbf{x})$. Next we pass this feature representation to the classifier to get the final prediction $\hat{y}= \arg\max_{y\in\mathcal{Y}^u}\mathcal{C}(\mathbf{f}_{test})[y]$

\vspace{-3pt}
\section{Experiments and Results}
\label{expt}

\begin{table}[t]
\centering
\scalebox{0.65}{
		\begin{tabular}{|c|c|c|c|c|c|c|c|}
		\toprule
		\multicolumn{2}{|c |}{\textbf{Method}} & \multicolumn{5}{c |}{\textbf{Target Domain}} &\\
		DG & ZSL & \textit{clipart} & \textit{infograph} & \textit{painting} & \textit{quickdraw} & \textit{sketch} & Avg.\\
		\midrule
                      \multirow{3}{*}{-}&       DEVISE~    & 20.1 &11.7  &17.6  &6.1   &16.7  &14.4 \\
                     &        ALE~    & 22.7&12.7  &20.2  &6.8  &18.5   &16.2 \\
                      &       SPNet~     &{26.0}  &16.9  & 23.8 & 8.2  & 21.8  &19.4 \\
                             
                             		\midrule
		
                             \multirow{3}{*}{DANN~}&DEVISE~   & 20.5&10.4&16.4&7.1&15.1&13.9 \\
                             
                             &ALE~     & 21.2&12.5&19.7&7.4&17.9&15.7 \\
         &SPNet~&25.9&15.8&24.1&8.4&21.3&19.1\\
		\midrule
		
                             \multirow{3}{*}{EpiFCR~}&DEVISE~    & 21.6& 13.9 &19.3  &7.3  &17.2   &15.9 \\
                             
                             &ALE~     & 23.2&  14.1&21.4  &7.8  &20.9   &17.5 \\
       &SPNet~ 
                               &{26.4} & 16.7  & 24.6  & 9.2 & 23.2   &20.0\\
        \midrule
        \multicolumn{2}{|c|}{Mixup-img-only} & 25.2 & 16.3 & 24.4 & 8.7 & 21.7 & 19.2 \\
        \multicolumn{2}{|c|}{Mixup-two-level} & 26.6 & 17 & 25.3 & 8.8 & 21.9 & 19.9 \\
        \multicolumn{2}{|c|}{CuMix  }  
                                     &\underline{27.6} &17.8 & 25.5  & 9.9  &22.6  & 20.7  \\
        \midrule
        \multicolumn{2}{|c|}{f-clsWGAN} &20.0 &13.3 &20.5 &6.6 &14.9 &15.1\\
        \multicolumn{2}{|c|}{CuMix + f-clsWGAN} &27.3 &\underline{17.9} &26.5 &11.2 &24.8 & 21.5\\
        \multicolumn{2}{|c|}{ROT + f-clsWGAN} &27.5 &17.4 &26.4 &11.4 &24.6 & 21.4\\ \midrule
        \multicolumn{2}{|c |}{ $\text{COCOA}_{AGG}$}    
                                     &27.6 &17.1  &25.7 & 11.8  &23.7  & 21.2 \\

        
        \multicolumn{2}{|c |}{$\textit{COCOA}_{ROT}$}    
                                     &\textbf{28.9} &\textbf{18.2}  & \underline{27.1}  & \textbf{13.1}  & \textbf{25.7}  & \textbf{22.6}  \\
        \bottomrule

		\end{tabular}}
		\vspace{-6pt}
		\caption{ \footnotesize Performance comparison with established baselines and state-of-art methods for ZSLDG problem setting  on benchmark DomainNet dataset. For fair comparison, all reported results follow the backbones, protocol and splits as established in CuMix. Best results are highlighted in bold and second best results are underlined.}
		\vspace{-8pt}
		\label{tab:domainnet-additional}
\end{table}
\vspace{-2pt}

\noindent \textbf{DomainNet Dataset:} DomainNet is the only established, diverse large-scale, coarse-grained dataset for ZSLDG \cite{dgzsl} with images belonging to 345 different categories divided into 6 different domains i.e \textit{painting}, \textit{real}, \textit{clipart}, \textit{infograph}, \textit{quickdraw} and \textit{sketch}. We follow the seen-unseen class training-testing splits and protocol as established in \cite{dgzsl} where we train on 5 domains and test on the left-out domain. Also, following \cite{dgzsl}, we use \textit{word2vec} representations \cite{word2vec} as the semantic representations for class labels. For feature extractor $f(.)$, we choose a Resnet-50 architecture similar to \cite{dgzsl}.


\vspace{3pt}
\noindent \textbf{Baselines.} We compare our approach with simpler baselines established in \cite{dgzsl} which include ZSL methods like \textit{SPNet} \cite{xian2019semantic}, \textit{DeViSE} \cite{devise}, \textit{ALE} \cite{akata2013label} and their coupling with well-known DG methods (DANN \cite{ganin2016domain}, EpiFCR \cite{li2019episodic}). We also compare with SOTA ZSLDG method, \textit{CuMix} \cite{dgzsl} as well as its variants: (1) \textit{Mixup-img-only} where mixup is done only at image level without curriculum; (2) \textit{Mixup-two-level} where mixup is applied at both feature and image level without curriculum.
We also establish Feature generation baselines which include f-clsWGAN \cite{zslgen0} (standard ZSL only approach) and its combination with visual backbone trained using CuMix \cite{dgzsl} methodology (\textit{CuMix + f-clsWGAN}) and self-supervised rotation \cite{RotNet} feature extraction method (\textit{ROT + f-clsWGAN}). 


\vspace{3pt}
\noindent \textbf{Results on DomainNet Benchmark.}
Table \ref{tab:domainnet-additional} shows the performance comparison of our method with the aforementioned baselines.
We observe that a simple classification-based feature extraction (using only $\mathcal{L_{AGG}}$) without any regularization when coupled with our generative module i.e $\textit{COCOA}_{AGG}$ is able to outperform the current state-of-the-art, CuMix \cite{dgzsl}.
In addition, when we use rotation prediction as a a regularizing auxiliary task \cite{RotNet} (referred as $\textit{COCOA}_{ROT}$), we observe that it achieves the best performance on all domains individually as well as on average (with significant improvement especially on hard domains like \emph{quickdraw} where there is large domain shift encountered at test-time), thus outperforming \cite{dgzsl} by a margin of about 2\% average across domains (which corresponds to a 10\% relative increase). We believe this is because the auxiliary task enables better representation learning. Also, we observe that combining generative ZSL baselines like f-clsWGAN \cite{zslgen0} with different visual backbones including CuMix results in inferior average performance when compared with our approach.


\vspace{3pt}




\noindent \textbf{Component-wise Ablation Study.}
Table \ref{tab_ablation} shows the component-wise performance for our method $COCOA_{ROT}$ on DomainNet dataset, following \cite{dgzsl}.
\vspace{-4pt}
\begin{itemize}[leftmargin=*]
\setlength\itemsep{-0.1em}
    \item $S1$ corresponds to the performance of the feature extractor $f(.)$ trained using  
    using rotation prediction as a regularizer (without generative stage 2).
    \item $S2$ corresponds to the performance achieved by learning a generator $G$ without using domain embeddings as an input. Specifically, this implies that the BatchNorm estimator networks $B_{gen}^{l}$ is given only the class-level semantic attribute as input i.e context vector $\mathbf{c} = \mathbf{a}^s_{y}$.
    \item $S3$ denotes the use of $S2$, with domain embeddings as an additional input in the context vector provided to $B_{gen}^{l}$ i.e $\mathbf{c} = [\mathbf{a}^s_{y},\mathbf{e}_{d}^{gen}]$, without the use of interpolated domain embeddings $\mathcal{E}_{interp}^{gen}$ (Eqn \ref{no-mix-inference}) and S4 denotes our complete model with the use of interpolated domain embeddings $\mathcal{E}_{interp}^{gen}$ to generate features. Note that $S2$, $S3$ and $S4$ follow the inference mechanism described in Sec \ref{inf}.
    
\end{itemize}
From Table \ref{tab_ablation}, we observe that $S4$ (proposed approach), which exploits the generative pipeline, semantic and domain embeddings as well as their mixing, achieves the best performance.
This corroborates our hypothesis that conditioning on both domain and semantic embeddings enables the classifier to discriminate between the distribution of features $\mathbf{f}$ better by encoding both domain-specific and class-level information in generated features $\hat{\mathbf{f}}$. Furthermore, training the final classifier on features generated by interpolating source domain embeddings (i.e $S4$)  further improves performance by alleviating the bias towards source domains.
\begin{table}
    \centering
    \scalebox{0.75}{
    \begin{tabular}{|c|c|c|c|c|c|c|}
    \toprule
    \textbf{Variant} & \textit{Clipart} & \textit{Infograph} & \textit{Painting} & \textit{Quickdraw} & \textit{Sketch} & Avg\\
    \toprule 
    $S1$ & 27.5	& 17.8 & 	25.4 &	9.5	7 & 22.5 &	20.54  \\
    $S2$ & 27.6	7 & 17.36 &	27.08 &	11.57 &	24.97	& 21.716  \\     
    $S3$ & 28.5	& 17.6	& 26.8 &	12.7 &	25.48 & 22.2  \\
    $S4$ & 28.9 & 18.2  & 27.1  & 13.1  & 25.7  & 22.6\\
    \hline
    \end{tabular}}
    \vspace{-2pt}
    \caption{\footnotesize Ablation study for different components of our framework on DomainNet dataset}
    \vspace{-14pt}
    \label{tab_ablation}
\end{table} 

\begin{figure}[!b]
  \vspace{-10pt}
    \includegraphics[width=1\linewidth]{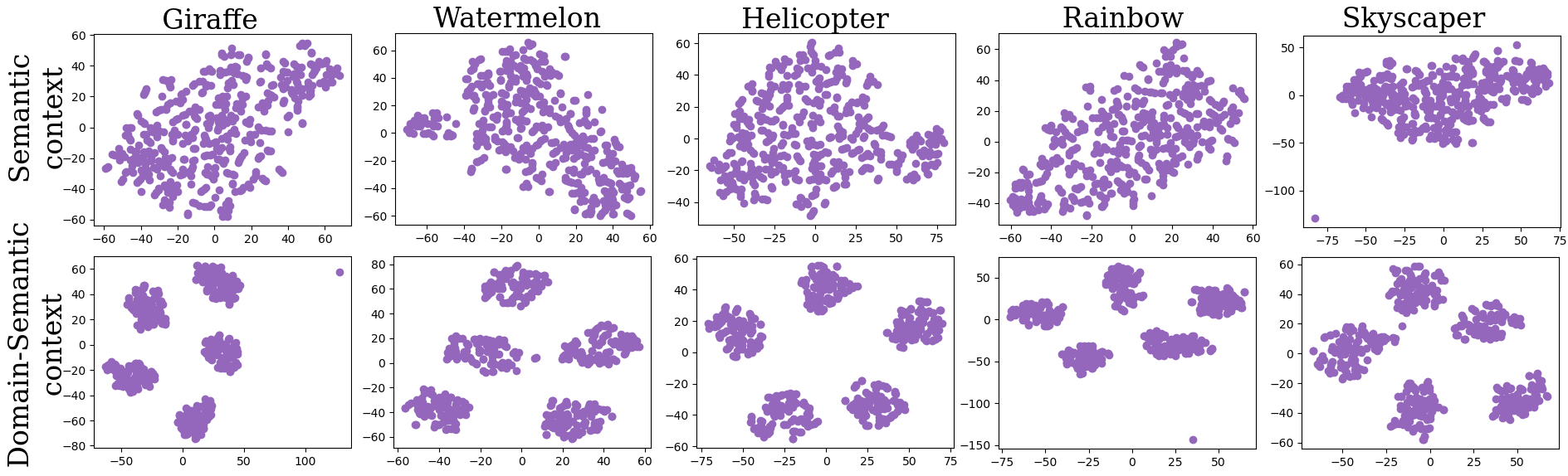}
    \caption{\footnotesize{Individual t-SNE visualization of synthesized image features by COCOA for randomly selected unseen classes (\textit{Giraffe}, \textit{Watermelon}, \textit{Helicopter}, \textit{Rainbow}, \textit{Skyscraper}) using only semantic context (Row 1) and both domain-semantic context (Row 2).
    }}
    \label{fig:tsne_cub}
\end{figure}



\noindent \textbf{Visualization of Generated Features.}
 We sample semantic attributes $\mathbf{a}_{y}^{u}$ for randomly selected unseen classes and use domain embeddings (learned end-to-end) of the five source domains (\textit{Real}, \textit{Infograph}, \textit{Quickdraw}, \textit{Clipart}, \textit{Sketch}) used during training. 
We individually visualize the features generated of each unseen class using only semantic embeddings/context i.e $\mathbf{c} = \mathbf{a}^u_{y}$ (Fig \ref{fig:tsne_cub}, Row 1) and concatenation of both semantic and domain embeddings/context i.e $\mathbf{c} = [\mathbf{a}^u_{y},\mathbf{e}_{d}^{gen}]$ (Fig \ref{fig:tsne_cub}, Row 2) when estimating the batchnorm parameters of the generator.
We notice that conditioning the batchnorm parameters on both semantic and domain context (Fig \ref{fig:tsne_cub}, Row 2) enables the model to better captures the modes of the original data distribution. It can be seen that the model can better retain domain-specific variance (associated with the five source domains) within a specific class cluster when both semantic and domain embeddings are used (Fig \ref{fig:tsne_cub}, Row 2),  compared to the case where only semantic context is used (Fig \ref{fig:tsne_cub}, Row 1) 
\vspace{-5pt}
\section{Conclusion}
\label{others}
\vspace{-3pt}
In this work, we propose a unified generative framework for the ZSLDG setting that uses context conditional batch-normalization to integrate class-level and domain-specific information into generated visual features, thereby enabling better generalization at test time. Through experiments, we demonstrate superior performance over established baselines and SOTA. Our proposed approach can be seamlessly integrated into other generative frameworks like VAEs, Flows, etc. which is left for future work.

\bibliographystyle{named}
\bibliography{ijcai21}

\begin{thebibliography}{}

\bibitem[\protect\citeauthoryear{Akata \bgroup \em et al.\egroup
  }{2013}]{akata2013label}
Zeynep Akata, Florent Perronnin, Zaid Harchaoui, and Cordelia Schmid.
\newblock Label-embedding for attribute-based classification.
\newblock In {\em Proceedings of the IEEE Conference on Computer Vision and
  Pattern Recognition}, pages 819--826, 2013.

\bibitem[\protect\citeauthoryear{Carlucci \bgroup \em et al.\egroup
  }{2019}]{jigsaw}
Fabio~Maria Carlucci, Antonio D'Innocente, Silvia Bucci, B.~Caputo, and
  T.~Tommasi.
\newblock Domain generalization by solving jigsaw puzzles.
\newblock {\em 2019 IEEE/CVF Conference on Computer Vision and Pattern
  Recognition (CVPR)}, pages 2224--2233, 2019.

\bibitem[\protect\citeauthoryear{Chandhok and Balasubramanian}{2021}]{taco}
Shivam Chandhok and V.~Balasubramanian.
\newblock Two-level adversarial visual-semantic coupling for generalized
  zero-shot learning.
\newblock {\em WACV}, 2021.

\bibitem[\protect\citeauthoryear{Chattopadhyay \bgroup \em et al.\egroup
  }{2020}]{BalanceSpecInv}
Prithvijit Chattopadhyay, Y.~Balaji, and Judy Hoffman.
\newblock Learning to balance specificity and invariance for in and out of
  domain generalization.
\newblock volume abs/2008.12839, 2020.

\bibitem[\protect\citeauthoryear{Felix \bgroup \em et al.\egroup
  }{2018}]{zslgen5}
Rafael Felix, Vijay~BG Kumar, Ian Reid, and Gustavo Carneiro.
\newblock Multi-modal cycle-consistent generalized zero-shot learning.
\newblock {\em In Proceedings of the European Conference on Computer Vision},
  2018.

\bibitem[\protect\citeauthoryear{Frome \bgroup \em et al.\egroup
  }{2013}]{devise}
Andrea Frome, Greg~S Corrado, Jon Shlens, Samy Bengio, Jeff Dean, Marc'Aurelio
  Ranzato, and Tomas Mikolov.
\newblock Devise: A deep visual-semantic embedding model.
\newblock In {\em Advances in neural information processing systems}, pages
  2121--2129, 2013.

\bibitem[\protect\citeauthoryear{Ganin \bgroup \em et al.\egroup
  }{2016}]{ganin2016domain}
Yaroslav Ganin, Evgeniya Ustinova, Hana Ajakan, Pascal Germain, Hugo
  Larochelle, Fran{\c{c}}ois Laviolette, Mario Marchand, and Victor Lempitsky.
\newblock Domain-adversarial training of neural networks.
\newblock {\em The Journal of Machine Learning Research}, 17(1):2096--2030,
  2016.

\bibitem[\protect\citeauthoryear{Ghifary \bgroup \em et al.\egroup
  }{2015}]{MTAE}
Muhammad Ghifary, W.~Kleijn, M.~Zhang, and D.~Balduzzi.
\newblock Domain generalization for object recognition with multi-task
  autoencoders.
\newblock {\em 2015 IEEE International Conference on Computer Vision (ICCV)},
  pages 2551--2559, 2015.

\bibitem[\protect\citeauthoryear{Gidaris \bgroup \em et al.\egroup
  }{2018}]{RotNet}
Spyros Gidaris, Praveer Singh, and Nikos Komodakis.
\newblock Unsupervised representation learning by predicting image rotations.
\newblock {\em ArXiv}, abs/1803.07728, 2018.

\bibitem[\protect\citeauthoryear{Huang \bgroup \em et al.\egroup
  }{2019}]{zslgen6}
He~Huang, Changhu Wang, Philip~S Yu, and Chang-Dong Wang.
\newblock Generative dual adversarial network for generalized zero-shot
  learning.
\newblock {\em CVPR}, 2019.

\bibitem[\protect\citeauthoryear{Keshari \bgroup \em et al.\egroup
  }{2020}]{zslgen7}
Rohit Keshari, Richa Singh, and Mayank Vatsa.
\newblock Generalized zero-shot learning via over-complete distribution.
\newblock {\em CVPR}, 2020.

\bibitem[\protect\citeauthoryear{Kodirov \bgroup \em et al.\egroup
  }{2015}]{semantic2vis2}
Elyor Kodirov, Tao Xiang, Zhenyong Fu, and S.~Gong.
\newblock Unsupervised domain adaptation for zero-shot learning.
\newblock {\em 2015 IEEE International Conference on Computer Vision (ICCV)},
  pages 2452--2460, 2015.

\bibitem[\protect\citeauthoryear{Li \bgroup \em et al.\egroup
  }{2017}]{basicdg4}
Da~Li, Yongxin Yang, Yi-Zhe Song, and Timothy~M. Hospedales.
\newblock Deeper, broader and artier domain generalization.
\newblock {\em 2017 IEEE International Conference on Computer Vision (ICCV)},
  pages 5543--5551, 2017.

\bibitem[\protect\citeauthoryear{Li \bgroup \em et al.\egroup
  }{2018a}]{metadg1}
Da~Li, Yongxin Yang, Yi-Zhe Song, and Timothy~M. Hospedales.
\newblock Learning to generalize: Meta-learning for domain generalization.
\newblock In {\em AAAI}, 2018.

\bibitem[\protect\citeauthoryear{Li \bgroup \em et al.\egroup }{2018b}]{DAFL}
Haoliang Li, Sinno~Jialin Pan, S.~Wang, and A.~Kot.
\newblock Domain generalization with adversarial feature learning.
\newblock {\em 2018 IEEE/CVF Conference on Computer Vision and Pattern
  Recognition}, pages 5400--5409, 2018.

\bibitem[\protect\citeauthoryear{Li \bgroup \em et al.\egroup
  }{2019a}]{metadg2}
Da~Li, J.~Zhang, Yongxin Yang, Cong Liu, Yi-Zhe Song, and Timothy~M.
  Hospedales.
\newblock Episodic training for domain generalization.
\newblock {\em 2019 IEEE/CVF International Conference on Computer Vision
  (ICCV)}, pages 1446--1455, 2019.

\bibitem[\protect\citeauthoryear{Li \bgroup \em et al.\egroup
  }{2019b}]{li2019episodic}
Da~Li, Jianshu Zhang, Yongxin Yang, Cong Liu, Yi-Zhe Song, and Timothy~M
  Hospedales.
\newblock Episodic training for domain generalization.
\newblock In {\em Proceedings of the IEEE International Conference on Computer
  Vision}, pages 1446--1455, 2019.

\bibitem[\protect\citeauthoryear{Mancini \bgroup \em et al.\egroup
  }{2020}]{dgzsl}
Massimiliano Mancini, Zeynep Akata, E.~Ricci, and Barbara Caputo.
\newblock Towards recognizing unseen categories in unseen domains.
\newblock In {\em ECCV}, 2020.

\bibitem[\protect\citeauthoryear{Maniyar \bgroup \em et al.\egroup
  }{2020}]{ZeroShotDG}
Udit Maniyar, K.~J. Joseph, A.~Deshmukh, {\"U}.~Dogan, and V.~Balasubramanian.
\newblock Zero-shot domain generalization.
\newblock {\em ArXiv}, abs/2008.07443, 2020.

\bibitem[\protect\citeauthoryear{Mikolov \bgroup \em et al.\egroup
  }{2013}]{word2vec}
Tomas Mikolov, Kai Chen, Greg~S. Corrado, and Jeffrey Dean.
\newblock Efficient estimation of word representations in vector space, 2013.

\bibitem[\protect\citeauthoryear{Mishra \bgroup \em et al.\egroup
  }{2018}]{zslgen4}
Ashish Mishra, Shiva~Krishna Reddy, Anurag Mittal, and Hema~A Murthy.
\newblock A generative model for zero shot learning using conditional
  variational autoencoders.
\newblock {\em CVPRW}, 2018.

\bibitem[\protect\citeauthoryear{Miyato and Koyama}{2018}]{miyato2018cgans}
Takeru Miyato and Masanori Koyama.
\newblock c{GAN}s with projection discriminator.
\newblock In {\em International Conference on Learning Representations}, 2018.

\bibitem[\protect\citeauthoryear{Muandet \bgroup \em et al.\egroup
  }{2013}]{basicdg2}
Krikamol Muandet, D.~Balduzzi, and B.~Sch{\"o}lkopf.
\newblock Domain generalization via invariant feature representation.
\newblock {\em ArXiv}, abs/1301.2115, 2013.

\bibitem[\protect\citeauthoryear{Narayan \bgroup \em et al.\egroup
  }{2020}]{tfvaegan}
Sanath Narayan, A.~Gupta, F.~Khan, Cees G.~M. Snoek, and L.~Shao.
\newblock Latent embedding feedback and discriminative features for zero-shot
  classification.
\newblock {\em ArXiv}, abs/2003.07833, 2020.

\bibitem[\protect\citeauthoryear{Ni \bgroup \em et al.\egroup }{2019}]{zslgen2}
Jian Ni, Shanghang Zhang, and Haiyong Xie.
\newblock Dual adversarial semantics-consistent network for generalized
  zero-shot learning.
\newblock {\em NeurIPS,}, 2019.

\bibitem[\protect\citeauthoryear{Reed \bgroup \em et al.\egroup
  }{2016}]{vis2semantic1}
Scott~E. Reed, Zeynep Akata, H.~Lee, and B.~Schiele.
\newblock Learning deep representations of fine-grained visual descriptions.
\newblock {\em 2016 IEEE Conference on Computer Vision and Pattern Recognition
  (CVPR)}, pages 49--58, 2016.

\bibitem[\protect\citeauthoryear{Schonfeld \bgroup \em et al.\egroup
  }{2019}]{zslgen3}
Edgar Schonfeld, Sayna Ebrahimi, Samarth Sinha, Trevor Darrell, and Zeynep
  Akata.
\newblock Generalized zero-and few-shot learning via aligned variational
  autoencoders.
\newblock {\em CVPR}, 2019.

\bibitem[\protect\citeauthoryear{Seo \bgroup \em et al.\egroup }{2020}]{BNE2}
Seonguk Seo, Yumin Suh, D.~Kim, Jongwoo Han, and B.~Han.
\newblock Learning to optimize domain specific normalization for domain
  generalization.
\newblock 2020.

\bibitem[\protect\citeauthoryear{Shen \bgroup \em et al.\egroup
  }{2020}]{normflows}
Yuming Shen, J.~Qin, and L.~Huang.
\newblock Invertible zero-shot recognition flows.
\newblock In {\em ECCV}, 2020.

\bibitem[\protect\citeauthoryear{Wan \bgroup \em et al.\egroup
  }{2019}]{semantic2vis1}
Ziyu Wan, Dongdong Chen, Y.~Li, Xingguang Yan, Junge Zhang, Y.~Yu, and Jing
  Liao.
\newblock Transductive zero-shot learning with visual structure constraint.
\newblock In {\em NeurIPS}, 2019.

\bibitem[\protect\citeauthoryear{Xian \bgroup \em et al.\egroup
  }{2018}]{zslgen0}
Yongqin Xian, Tobias Lorenz, Bernt Schiele, , and Zeynep Akata.
\newblock Feature generating networks for zero-shot learning.
\newblock {\em CVPR}, 2018.

\bibitem[\protect\citeauthoryear{Xian \bgroup \em et al.\egroup
  }{2019}]{xian2019semantic}
Yongqin Xian, Subhabrata Choudhury, Yang He, Bernt Schiele, and Zeynep Akata.
\newblock Semantic projection network for zero-and few-label semantic
  segmentation.
\newblock In {\em Proceedings of the IEEE Conference on Computer Vision and
  Pattern Recognition}, pages 8256--8265, 2019.

\bibitem[\protect\citeauthoryear{Xu \bgroup \em et al.\egroup
  }{2014}]{basicdg3}
Zheng Xu, W.~Li, Li~Niu, and Dong Xu.
\newblock Exploiting low-rank structure from latent domains for domain
  generalization.
\newblock In {\em ECCV}, 2014.

\bibitem[\protect\citeauthoryear{Yang and Gao}{2013}]{basicdg1}
P.~Yang and Wei Gao.
\newblock Multi-view discriminant transfer learning.
\newblock In {\em IJCAI}, 2013.

\bibitem[\protect\citeauthoryear{Zhang \bgroup \em et al.\egroup
  }{2017}]{semantic2vis3}
L.~Zhang, Tao Xiang, and S.~Gong.
\newblock Learning a deep embedding model for zero-shot learning.
\newblock {\em 2017 IEEE Conference on Computer Vision and Pattern Recognition
  (CVPR)}, pages 3010--3019, 2017.

\end{thebibliography}

\end{document}